\documentclass[11pt]{article}

\usepackage[preprint]{acl}

\usepackage{times}
\usepackage{latexsym}
\usepackage[dvipsnames]{xcolor}
\usepackage[T1]{fontenc}
\usepackage[utf8]{inputenc}

\usepackage{microtype}
\usepackage[most]{tcolorbox}
\usepackage{epigraph}
\usepackage{pifont}
\newcommand{\cmark}{\text{\ding{51}}}
\newcommand{\xmark}{\text{\ding{55}}}
\usepackage{inconsolata}
\usepackage{tabularx, booktabs}
\usepackage{booktabs}
\usepackage{siunitx}
\usepackage[table]{xcolor}
\usepackage{colortbl} % optional; only needed if you use \rowcolor/\rowcolors
\usepackage{amsmath,amssymb,amsfonts,amsthm}
\usepackage{graphicx}
\usepackage{fontawesome5}
\usepackage{subcaption}
\usepackage{tikz}
\definecolor{peachbadge}{RGB}{248,224,203}
\newcommand{\cnum}[1]{%
  \begingroup
  \tikz[baseline=(c.base)]{\node[circle, fill=peachbadge, text=black,
        inner sep=1.2pt, font=\scriptsize\bfseries] (c) {#1};}%
  \endgroup
}
\usepackage{xspace}
\makeatletter
\DeclareRobustCommand\onedot{\futurelet\@let@token\@onedot}
\def\@onedot{\ifx\@let@token.\else.\null\fi\xspace}

\makeatother

\let\oldcolorbox\colorbox

\renewcommand{\colorbox}[2]{%
  \begingroup
  \setlength{\fboxsep}{0pt}% no extra vertical padding
  \oldcolorbox{#1}{\strut #2}%
  \endgroup
}

\newcommand{\EN}{\textcolor{blue!70!black}{\textbf{English}}}
\newcommand{\FR}{\textcolor{teal!70!black}{\textbf{French}}}
\newcommand{\ZH}{\textcolor{purple!70!black}{\textbf{Mandarin}}}
\newcommand{\AR}{\textcolor{orange!80!black}{\textbf{Arabic}}}
\newcommand{\NL}{\textcolor{green!50!black}{\textbf{Dutch}}}
\newcommand{\DE}{\textcolor{brown!80!black}{\textbf{German}}}
\newcommand{\IT}{\textcolor{red!70!black}{\textbf{Italian}}}
\newcommand{\ES}{\textcolor{violet!70!black}{\textbf{Spanish}}}

\definecolor{lightcyan}{HTML}{ADD8E6}
\definecolor{softyellow}{HTML}{FFF2A8}

\usepackage[capitalize]{cleveref}
\crefname{figure}{Fig.}{Figs.}
\Crefname{figure}{Figure}{Figures}
\crefname{section}{Sec.}{Secs.}
\Crefname{section}{Section}{Sections}
\crefname{table}{Tab.}{Tabs.}
\Crefname{table}{Table}{Tables}

\title{Vision-Language Models are Fragile Multilingual Associators}

\author{
  \textbf{Ritabrata Chakraborty\textsuperscript{1,4,5}},
  \textbf{Rajatsubhra Chakraborty\textsuperscript{2}},
  \\
  \textbf{Shivakumara Palaiahnakote\textsuperscript{3}},
  \textbf{Angelo Cangelosi\textsuperscript{4}},
  \textbf{Umapada Pal\textsuperscript{5}}
  \\
  \\
  \textsuperscript{1}Manipal University Jaipur, India \quad
  \textsuperscript{2}University of North Carolina Charlotte, USA
  \\
  \textsuperscript{3}University of Salford, UK \quad
  \textsuperscript{4}University of Manchester, UK
  \\
  \textsuperscript{5}Indian Statistical Institute Kolkata, India
  \\
  {\small \href{https://ritabrata04.github.io/m2bind/}
  {\faIcon{globe}\ \texttt{https://ritabrata04.github.io/m2bind/}}
  }
}

\begin{document}
\maketitle
\begin{abstract}

Vision-language models must associate visual entities with textual attributes. Whether these associations or concept bindings remain stable when the language of the input changes is unexplored. We introduce M$^2$BIND, a benchmark varying the language of the context and query across multiple languages. We evaluate binding both extrinsically through task performance metrics and intrinsically through causal interventions. We find that binding is not language-invariant: cross-family and cross-script settings trigger significant binding collapse, with the model's internal binding computation shifting to later layers and losing causal strength. Closely related languages preserve associations comparatively better. In a broader sense, our findings indicate how VLMs deployed globally in multilingual settings cannot be assumed to maintain the same association quality observed in monolingual evaluation.

\end{abstract}

\section{Introduction}
\label{sec:introduction}

\begin{figure}[t!]
    \centering
    \includegraphics[width=\linewidth]{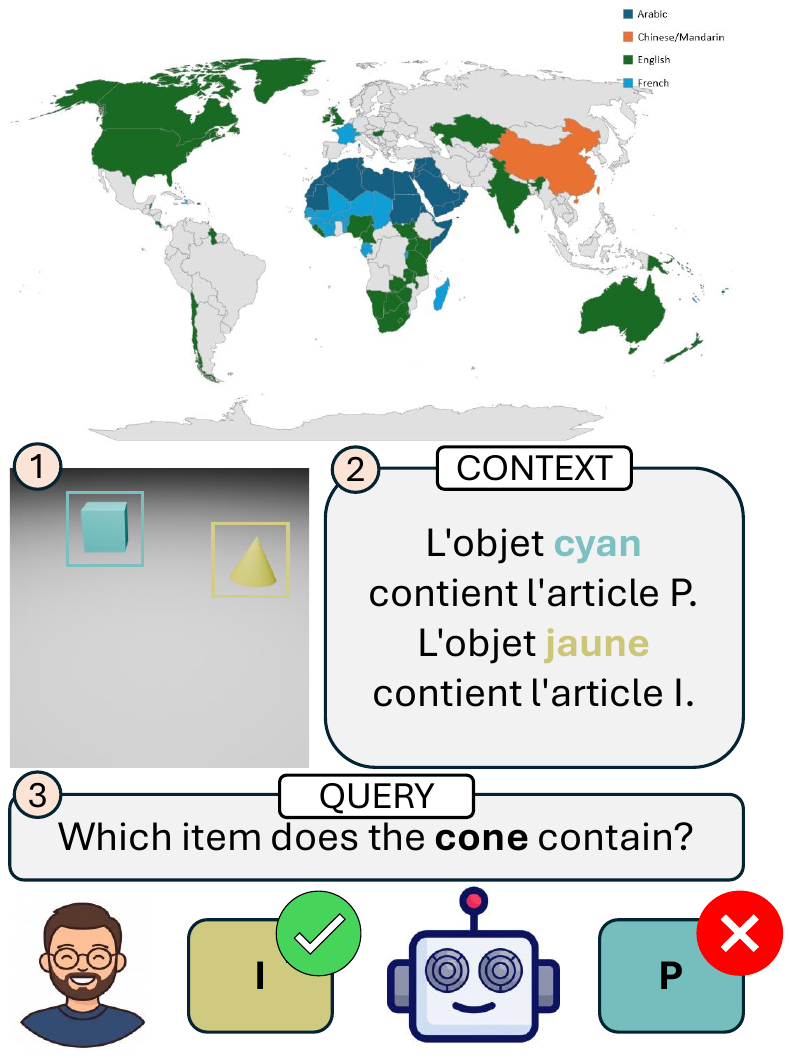}
    \caption{\textbf{Concept binding across languages.}
    \textbf{(Top)} The world map situates the languages in our study across
    continents, families, and scripts. \textbf{(Bottom)} A \FR{} speaker who also
    reads \EN{} effortlessly maps \protect\colorbox{softyellow}{\strut jaune} onto
    the same concept as \protect\colorbox{softyellow}{\strut yellow}, and so links
    the 1 image, the 2 \FR{} context, and the 3 \EN{} query to answer
    \textbf{cone}\,$\rightarrow$\,I (\cmark). VLMs frequently fail this
    association (\xmark). We investigate whether such bindings survive when the
    context and query languages differ. 
    % Best viewed in color.
    }
    \label{fig:teaser}
    \vspace{-7pt}
\end{figure}

A reader who speaks \FR{} and \EN{} does not pause to translate when informed about something \protect\colorbox{softyellow}{\strut ``yellow"} and asked about something \protect\colorbox{softyellow}{\strut ``jaune"}. Humans fuse what we see with what we read into a single concept, and we do so without effort no matter which of the world's languages the words arrive in (\cref{fig:teaser}) \cite{bonner2021object, xie2017semantic}. These \underline{multilingual associations} come innately with our understanding of languages and how they describe (here, in the visual sense) the world around us.

Vision-language models (VLMs) \cite{openai2024gpt4ocard, gemini} now operate in exactly these settings, answering questions about images for users spread across the languages and scripts of the map in \cref{fig:teaser}. The languages depicted (top) are chosen to span multiple scripts and language families across several continents, so our findings speak to globally diverse deployment rather than to any single region. If the concept associations a VLM forms internally are tethered to the surface form of \EN{}, then every non-\EN{} user inherits a silent degradation that monolingual evaluation never exposes. 
% This is the question we study: \textbf{\textit{do a VLM's internal concept associations survive a change of language?}}
Such associations are governed by \emph{binding}, which in its simplest causal form asks : \texttt{given X:Y and Y:Z, can the model infer X:Z?} (see \cref{eq:chain}). \citet{feng2024how} showed that language models attach latent Binding IDs to entities, representing this structure in their activations without it ever appearing in the prompt; \citet{darshanavlm} extended the result to VLMs, where image tokens bind to their textual references. We subject this in-context binding to a novel test, where we establish the binding in one language and retrieve it in another. We organize our study around three questions:

\begin{tcolorbox}[
colback=green!8,
colframe=green!60!black,
boxrule=0.8pt,
arc=3pt,
left=4pt,
right=4pt,
top=4pt,
bottom=4pt
]
\textbf{RQ1:} Do visual-textual associations survive across languages?

% \vspace{2.5pt}

\textbf{RQ2:} What inside the VLM drives their degradation?

% \vspace{2.5pt}

\textbf{RQ3:} Does linguistic closeness aid binding transfer?
\end{tcolorbox}

\paragraph{Contributions.} We (i) introduce \textsc{M$^2$Bind}, a task and benchmark that varies the context and query language independently over eight languages along two axes of linguistic distance; (ii) evaluate binding both extrinsically and intrinsically; (iii) trace the behaviour to linguistic distance, model tokenizer and decoder computations. 
To our knowledge, this is the first work of its kind.

% \noindent Across eight languages we find that binding holds to be stable within language families but collapses across scripts, with the model's causal binding computation drifting to later layers and weakening as linguistic distance grows. 
\cref{sec:m2bind} defines our task, data, and metrics; \cref{sec:results} discusses results and  \cref{sec:conc} concludes the paper.
\section{Methodology: M$^{2}$BIND}
\label{sec:m2bind}

\paragraph{Task.}We introduce the task of \textbf{\underline{M}ultilingual \underline{M}ultimodal \underline{BIND}ing}: for a visual scene and a textual context that assigns attributes to objects in language $\ell_{\mathrm{ctx}}$, can a VLM retrieve the correct association when queried in a different language $\ell_q$?

\paragraph{Problem formulation.}(\cref{fig:teaser}) A scene $s=(v,\mathcal{O})$ consists of an image \cnum{1} $v$ and a set of objects $\mathcal{O}=\{o_1,o_2\}$. Each object $o_j$ is characterised by a shape $\sigma_j\in\Sigma$ and a colour $\gamma_j\in\Gamma$, with $\Sigma=\{\textrm{cone},\textrm{cube},\textrm{cylinder},\textrm{sphere}\}$ and $\Gamma=\{\textrm{red},\textrm{blue},\textrm{green},\textrm{yellow},\textrm{cyan},\textrm{purple}\}$. A textual context \cnum{2} $c$ refers to each object by its colour and assigns it an item symbol $\iota_j\in\mathcal{I}$, thereby defining a colour-to-item binding $b:\gamma_j\mapsto\iota_j$. A query \cnum{3} $q$ singles out a target object by its shape $\sigma_\star$ and asks for the item it contains. Producing the correct answer requires the model to compose two associations,
\begin{equation}
\sigma_\star \;\xrightarrow[\;v\;]{\textrm{ground}}\; \gamma_\star \;\xrightarrow[\;c\;]{\textrm{bind}}\; \iota_\star = b(\gamma_\star),
\label{eq:chain}
\end{equation}
first \underline{grounding} the queried shape to a colour through the image, then resolving that colour to its \underline{bound} item through the context (\cref{eq:chain}). Crucially, associations must be carried by \emph{latent binding variables} in the model's activations \citep{feng2024how,darshanavlm}.

\paragraph{Linguistic conditioning.}
We render the context in a language $\ell_{\mathrm{ctx}}$ and the query in a language $\ell_q$ while holding the image $v$ fixed, so an instance is the triple $x=(v,\,c^{\ell_{\mathrm{ctx}}},\,q^{\ell_q})$. 
Because the image, the underlying scene, and the gold item $\iota_\star$ are identical across all conditions, any change in the model's response isolates the effect of \emph{language} on binding rather than of perception or task difficulty.

\paragraph{Data and Setup.}
We build on the Shapes binding task of \citet{darshanavlm}, in which each instance pairs a Blender-rendered image of two 3D objects with a context that binds each object's colour to an item symbol and a query that targets one object by its shape. We extend the task by independently varying $\ell_{\mathrm{ctx}}$ and $\ell_q$ over $8$ languages selected along two axes of linguistic distance\footnote{We discuss linguistic choices in \cref{app:lang_details}.}: \textbf{(i)~cross-family}: \EN{}, \FR{}, \ZH{}, \AR{}, spanning four families and three scripts \citep{asher2018atlas, littell2017uriel}; and \textbf{(ii)~within-family}, comprising a \textbf{Germanic} cluster (\EN{}, \NL{}, \DE{}) and a \textbf{Romance} cluster (\FR{}, \IT{}, \ES{}) that share the Latin script but differ in genealogical proximity \citep{Campbell_Grondona_2008, gooskens2018mutual}\footnote{Combinations of these languages yield 16 arrangements for cross-family and 9 for within-family.}.

\paragraph{Model.}
We evaluate LLaVA-1.5-OV-7B \citep{an2025llavaonevision15fullyopenframework} a SigLIP vision encoder, an MLP projector, and Qwen2 decoder following \citet{darshanavlm} as a representative open-weight VLM. Given an instance $x$, the model returns a conditional distribution over candidate items at the answer position under greedy decoding. All reported values are averaged over $3$ runs on a single NVIDIA A40 (48GB) GPU.
\paragraph{Sequence score.}
We compare the two candidate items $\mathcal{C}=\{\iota_\star,\iota_{\mathrm{swap}}\}$, where $\iota_\star$ is bound to the queried object and $\iota_{\mathrm{swap}}$ to the distractor. For instance $x_i$, we score a candidate $I$ by the log-probability the model assigns to its token sequence $y_{1:m(I)}(I)$ at the answer position:
\begin{equation}
S_i(I)=\sum_{t=1}^{m(I)}\log p_\theta\!\bigl(y_t(I)\mid x_i,\,y_{<t}(I)\bigr).
\label{eq:score}
\end{equation}

\paragraph{Accuracy and Factorization Margin.}
\textbf{Accuracy} is the fraction of instances for which $S_i(\iota_\star)>S_i(\iota_{\mathrm{swap}})$ under the score of \cref{eq:score}. Because accuracy saturates before binding fully degrades, we additionally report the \textbf{Factorization Margin} (FM), the mean score separation between the correct and swapped items \citep{feng2024how}:
\begin{equation}
\mathrm{FM}=\frac{1}{N}\sum_{i=1}^{N}\bigl[S_i(\iota_{\star,i})-S_i(\iota_{\mathrm{swap},i})\bigr].
\label{eq:fm}
\end{equation}
A large positive FM indicates that the model cleanly separates the bound item from the distractor, whereas $\mathrm{FM}\!\to\!0$ signals a collapsed binding even when accuracy remains high.

\paragraph{Causal localization.}
Accuracy and FM characterise binding only at the output. To locate \emph{where} binding is consolidated inside the decoder, we apply an interchange intervention \citep{meng2022locating, vig2020causal}. Let $h_{k,j}$ denote the layer-$k$ hidden state at the token position of object $o_j$. We re-run the forward pass with the two objects' states interchanged, $\mathrm{do}(h_{k,1}\!\leftrightarrow\!h_{k,2})$, and define the \textbf{Intervention Causal Effect} (ICE) as the resulting drop in the correct-item score, averaged over instances:
\begin{equation}
\mathrm{ICE}_k=\mathbb{E}_i\!\Bigl[S_i(\iota_\star)-S_i\!\bigl(\iota_\star\mid \mathrm{do}(h_{k,1}\!\leftrightarrow\!h_{k,2})\bigr)\Bigr].
\label{eq:ice}
\end{equation}
A large positive $\mathrm{ICE}_k$ means that disrupting the layer-$k$ representation of the two objects sharply lowers the score of the correct item, i.e.\ binding-relevant information is actively represented at layer $k$. This gives us a profile of where the associations are made.
% The layer-wise profile $\{\mathrm{ICE}_k\}_k$ thus provides a causal footprint of where the model commits to its entity--attribute associations.
\section{Results and Discussions}
\label{sec:results}

\begin{table}[h]
\centering
\footnotesize
\setlength{\tabcolsep}{2.0pt}
\renewcommand{\arraystretch}{1.1}
\definecolor{hlblue}{RGB}{225,240,255}
\begin{tabular}{@{}l*{4}{>{\centering\arraybackslash}p{0.15\columnwidth}}@{}}
\toprule
\textbf{$\ell_{\mathrm{ctx}}$ $\downarrow$ / $\ell_{\mathrm{q}}$ $\rightarrow$} &
\textcolor{blue!70!black}{\textbf{En}} &
\textcolor{teal!70!black}{\textbf{Fr}} &
\textcolor{purple!70!black}{\textbf{Zh}} &
\textcolor{orange!80!black}{\textbf{Ar}} \\
\midrule
\textcolor{blue!70!black}{\textbf{En}} & 0.995 / 5.45 & 0.990 / 5.20 & 0.955 / 3.85 & 0.940 / 3.40 \\
\textcolor{teal!70!black}{\textbf{Fr}} & 0.990 / 5.10 & 0.995 / 5.25 & 0.945 / 3.60 & \textcolor{red}{0.930 / 3.25} \\
\textcolor{purple!70!black}{\textbf{Zh}} & 0.965 / 4.10 & 0.955 / 3.85 & 0.985 / 4.95 & \textcolor{red}{0.900 / 2.80} \\
\textcolor{orange!80!black}{\textbf{Ar}} & 0.950 / 3.75 & 0.940 / 3.50 & \textcolor{red}{0.905 / 2.85} & 0.975 / 4.55 \\
\bottomrule
\end{tabular}
\vspace{2pt}
\caption{\textbf{Cross-family VLM associations.} Cross-lingual concept binding over \textcolor{blue!70!black}{\textbf{En}}, \textcolor{teal!70!black}{\textbf{Fr}}, \textcolor{purple!70!black}{\textbf{Zh}}, and \textcolor{orange!80!black}{\textbf{Ar}}; rows are context (ctx), columns are query (q). Each cell reports \emph{Acc.\ / FM}; for both, higher ($\uparrow$) is better. Worst cases are \textcolor{red}{highlighted}.}
\label{tab:pairwise}
\vspace{-6pt}
\end{table}
\paragraph{(RQ1) Binding across language families.}
Binding strength is far from language-invariant: FM falls by 2.65, from 5.45 for monolingual \EN{} to 2.80 for the \ZH{}--\AR{} cross-script pair, less than half its best monolingual value. \cref{tab:pairwise} reports accuracy and FM for all 16 context--query combinations over \EN{}, \FR{}, \ZH{}, and \AR{}. The monolingual diagonal ($\ell_{\mathrm{ctx}}{=}\ell_{\mathrm{q}}$) sets each language's upper bound: \EN{} is strongest at 5.45, while the two non-Latin scripts sit lowest even on their own diagonal (\ZH{} 4.95, \AR{} 4.55), so script already shapes binding before any language mixing. Cross-lingually, FM tracks typological distance. The close \EN{}--\FR{} pair stays near its monolingual values, whereas mixing a Latin with a non-Latin script collapses FM to roughly 3.2--4.1, and the two non-Latin scripts together reach the 2.80 floor noted above with accuracy at its $\sim$0.90 floor. Degradation is also \underline{asymmetric}: across the four Latin/non-Latin pairs, the non-Latin language on the query side is consistently harder than on the context side by about 0.25 FM (e.g.\ \EN{}$\to$\AR{} 3.40 vs.\ \AR{}$\to$\EN{} 3.75). This matches the Matrix Language Frame account of code-switching \citep{myers2001matrix}, where $\ell_{\mathrm{ctx}}$ builds the binding scaffold and $\ell_{\mathrm{q}}$ must access it: access is more fragile than construction once the frames diverge. The trend is consistent with multilingual NLP \citep{lauscher-etal-2020-zero, pires-etal-2019-multilingual}.

\begin{table}[t]
\centering
\small
\setlength{\tabcolsep}{6pt}
\renewcommand{\arraystretch}{1}
\begin{tabular}{@{}l S[table-format=4.0] S[table-format=2.0] S[table-format=1.2]@{}}
\toprule
\textbf{Lang} &
{\textbf{Mean tokens}~$\downarrow$} &
{\textbf{Trunc.}~(\%)~$\downarrow$} &
{\textbf{FM drop}~$\downarrow$} \\
\midrule
\textcolor{blue!70!black}{\textbf{En}} & 1450 & 0  & 0.00 \\
\textcolor{teal!70!black}{\textbf{Fr}} & 1520 & 0  & 0.20 \\
\textcolor{purple!70!black}{\textbf{Zh}} & 1720 & 5  & 1.70 \\
\textcolor{orange!80!black}{\textbf{Ar}} & 1850 & 12 & 2.55 \\
\bottomrule
\end{tabular}
\caption{\textbf{Impact of tokenization.} Values reported are for monolingual setting. We report mean tokens of the image+context+query, truncation (Trunc.) against the context limit, and the FM drop relative to \EN{}. For all metrics, lower ($\downarrow$) is better.}
\vspace{-12pt}
\label{tab:tokenization}
\end{table}

\paragraph{(RQ2) Tokenization.}
Which component drives this fragility? Tokenization in particular plays a significant role since the same sentence can use a varying amount of tokens based on the language\footnote{This is referred to as the \textit{fertility} of a tokenizer \cite{rust-etal-2021-good}.}. In the monolingual setting (\cref{tab:tokenization}), \AR{} needs 27.6\% more tokens than \EN{}, truncates 12\% of prompts at the context limit, and loses 2.55 FM, with no cross-lingual transfer involved. Binding thus weakens within a single language from token budget alone, implicating tokenizer unfairness \citep{petrov2023token_unfairness} and supporting \citet{rust-etal-2021-good} on tokenizer quality and multilingual performance.

\begin{figure}[t]
    \centering
    \includegraphics[width=1\linewidth]{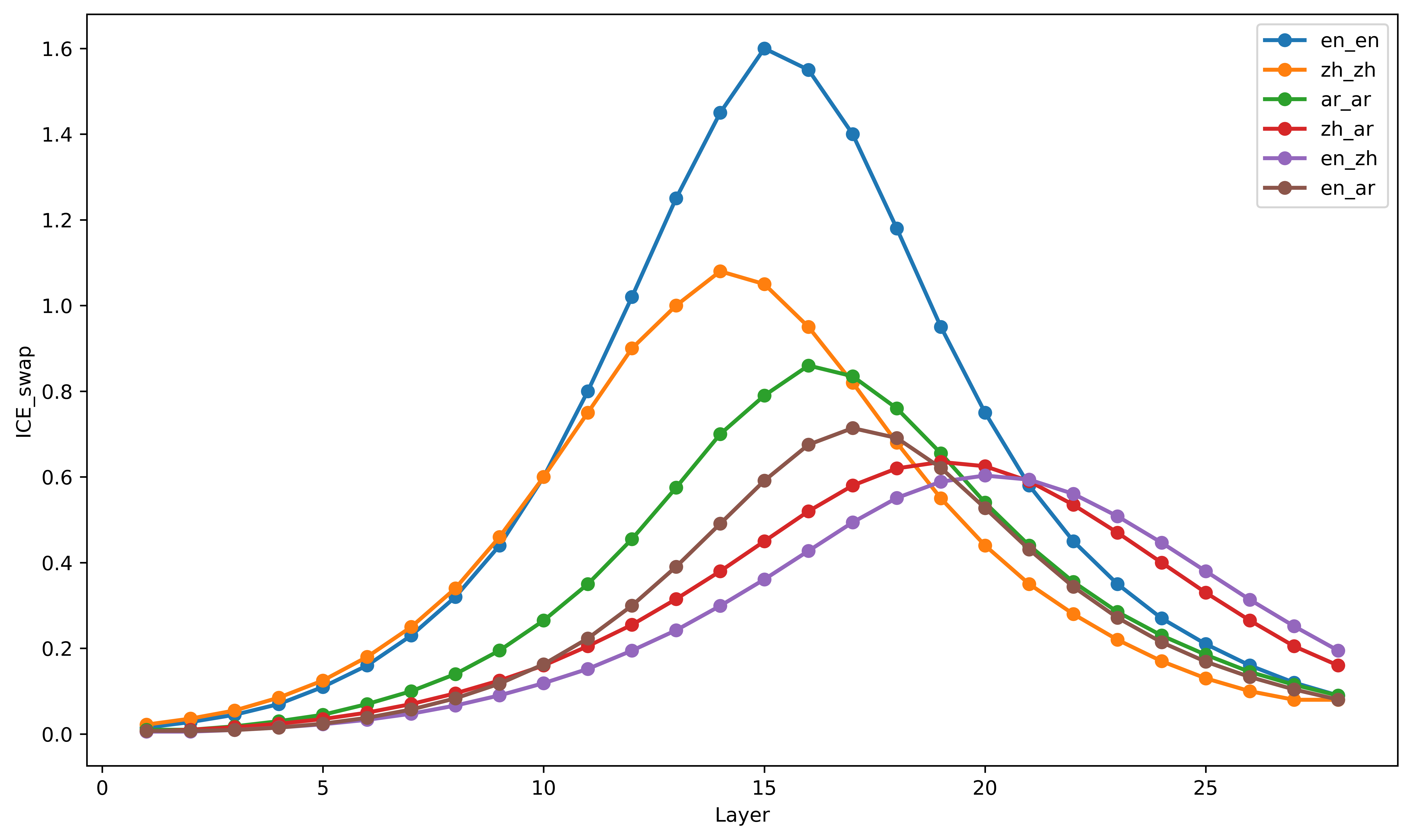}
    \caption{\textbf{Causal interventions for multilingual binding.} The $y$-axis is the change in log-probability (ICE) after intervention; the $x$-axis is decoder layer index. A higher peak implies stronger association strength at those layers. We show monolingual (\EN{}, \ZH{}, \AR{}) and cross-lingual combinations. Higher is better.}
    \label{fig:ICE}
    \vspace{-6pt}
\end{figure}

\paragraph{(RQ2) Binding in  VLM.}
While \cref{tab:tokenization} explains impact on binding before actual associations, we ask how the internal representations of the model actually perform multilingual association. \cref{fig:ICE} provides some insights. All curves are unimodal, depicting a localized region of layers where multilingual association is observed the strongest. For monolingual cases all curves peak around the \underline{middle layers} ($\sim$15), indicating that the model has already committed to the correct association before the final generation layers. Cross-lingual instances undergo a shift towards the right, peaking around the \underline{late layers} of the decoder ($\sim$20). This points to extra computation reconciling the context- and query-language representations. Interestingly, binding is reduced for cross-lingual cases as compared to monolingual curves. The combination of \ZH{}-\AR{}, which produced the weakest FM in \cref{tab:pairwise}, also produces the flattest and weakest curve in \cref{fig:ICE}. This means that even after extra computations in the decoder, the binding is less causally separable , i.e, the swap intervention has less effect because the two items' representations are less distinct.

% ── Within-family binding tables ──
\begin{table}[h]
\centering
\footnotesize
\setlength{\tabcolsep}{2.0pt}
\renewcommand{\arraystretch}{1.15}
\begin{subtable}[t]{\linewidth}
\centering
\begin{tabular}{@{}l*{3}{>{\centering\arraybackslash}p{0.18\columnwidth}}@{}}
\toprule
\textbf{L$_{\mathrm{ctx}}$ $\downarrow$ / L$_{\mathrm{q}}$ $\rightarrow$} &
\EN{} & \NL{} & \DE{} \\
\midrule
\EN{} & 0.995 / 5.45 & 0.990 / 5.30 & 0.985 / 5.10 \\
\NL{} & 0.990 / 5.25 & 0.990 / 5.20 & \textcolor{red}{0.985 / 5.05} \\
\DE{} & \textcolor{red}{0.985 / 5.00} & \textcolor{red}{0.985 / 5.00} & 0.985 / 5.05 \\
\bottomrule
\end{tabular}
\caption{}
\label{tab:germanic}
\end{subtable}
\vspace{2pt}
\begin{subtable}[t]{\linewidth}
\centering
\begin{tabular}{@{}l*{3}{>{\centering\arraybackslash}p{0.18\columnwidth}}@{}}
\toprule
\textbf{L$_{\mathrm{ctx}}$ $\downarrow$ / L$_{\mathrm{q}}$ $\rightarrow$} &
\FR{} & \IT{} & \ES{} \\
\midrule
\FR{} & 0.995 / 5.25 & 0.990 / 5.15 & 0.990 / 5.05 \\
\IT{} & 0.980 / 5.10 & 0.990 / 5.15 & \textcolor{red}{0.985 / 5.00} \\
\ES{} & \textcolor{red}{0.985 / 5.00} & \textcolor{red}{0.985 / 5.00} & 0.990 / 5.10 \\
\bottomrule
\end{tabular}
\caption{}
\label{tab:romance}
\end{subtable}
\vspace{-2pt}
\caption{\textbf{Within-family binding for Germanic and Romance families.} Worst cases are \textcolor{red}{highlighted}.}
\label{tab:within_family}
\vspace{-10pt}
\end{table}

\paragraph{(RQ3) Binding within families.}
Instead of languages that do not seemingly overlap in mutual intelligibility, how do VLMs behave when the languages are \underline{closer} to each other\footnote{Consider a native \NL{} speaker, they might have some understanding upon encountering a query in \DE{}.}?  \cref{tab:within_family} shows concept binding for the two language families mentioned in \cref{sec:m2bind}. Overall, every within-family FM stays above 5, the worst drop being of 0.45 from \DE{} to other languages. In terms of accuracy, within-family excels with a lower bound around 0.98, as compared to a floor of 0.90 for cross-family interactions in \cref{tab:pairwise}. \cref{tab:germanic} shows results for the Germanic family. \EN{}-\NL{} is akin to \EN{} in its monolingual setting, closer than \EN{}-\FR{}, despite both being non-English languages. Interestingly, \DE{}-\NL{} pairs perform really well, generalizing beyond \EN{}. \cref{tab:romance} shows results for Romance family, where \IT{} demonstrates to be better than \ES{} as ctx or q. \ES{}-\IT{} as a pair shows an FM of 5, showing associations work within the family, without the need of \FR{} as an anchor language, supporting \citet{conneau-etal-2020-unsupervised}.

\section{Conclusion} 
\label{sec:conc}
We present M²BIND, a benchmark that varies the context and query language independently to test whether VLMs maintain entity-attribute bindings across languages. Binding dissociates with linguistic distance: distant languages and non-Latin scripts collapse both binding strength and its causal localization, so near-ceiling task accuracy is not a reliable predictor of cross-lingual binding fidelity. Language-invariant grounding, effortless for humans, remains unsolved for current VLMs.

\section{Limitations}
\label{sec:limitations}
The Shapes task uses procedurally generated Blender images; we use this since our interest was to show a simple task where multilingual associations are not performed properly. Still, noticing an even further degradation of this concept binding for real images remains, beyond two objects or simple shapes. Further we show our work primarily on a LLaVA VLM, along with additional results on Qwen2.5 VL in Appendix \cref{app:qwen}. A similar look into other commercial models could potentially illustrate our generalizations further.
\appendix
\section{Related Works}
\label{app:related}
 
\paragraph{Multilingual vision-language models.}
Vision-language models have advanced rapidly \citep{openai2024gpt4ocard,gemini}, yet most are trained on predominantly English data and lose accuracy on non-English input \citep{geigle2025centurio}. This gap has driven a wave of multilingual and multicultural systems and evaluations. On the modelling side, efforts such as Centurio \citep{geigle2025centurio} and Pangea \citep{yue2025pangea} study how language coverage and data mixture shape cross-lingual ability while preserving English performance. On the evaluation side, cross-lingual visual question answering benchmarks such as xGQA \citep{pfeiffer2022xgqa} and culturally grounded suites such as CVQA \citep{romero2024cvqa} measure how well models transfer across languages and cultures. A recent survey catalogues 31 models and 21 benchmarks and identifies a persistent tension between language neutrality and faithful cross-lingual behaviour \citep{manea2025survey}. These efforts share a common lens, namely they measure multilingual ability through end-task accuracy. We instead probe an internal property, the stability of the entity--attribute binding itself, and show that aggregate accuracy can stay near ceiling while binding strength degrades, which makes downstream scores an incomplete diagnostic for multilingual deployment.
 
\paragraph{Language for cross-modal understanding.}
The question of whether the language one uses shapes how one perceives the world predates modern NLP, originating in the linguistic relativity tradition \citep{whorf1956language, kay1984what}. A long line of psycholinguistic work has made this concrete in the visual domain, the same setting VLMs operate in. Speakers whose language lexicalises a colour distinction discriminate those colours faster, as shown for Russian light and dark blue \citep{winawer2007russian} and Greek blues \citep{thierry2009unconscious}, with category effects that are lateralised to the language-dominant hemisphere \citep{gilbert2006whorf} and that appear even pre-attentively\citep{roberson2000color, regier2009language}. Beyond colour, the language of thought has been argued to influence conceptions of time and other abstract domains \citep{boroditsky2001does}. Complementarily, neuroscientific evidence indicates that human object representations themselves reflect the co-occurrence statistics of vision and language \citep{bonner2021object} and that congruent visual and verbal signals are integrated during perception \citep{xie2017semantic}. Together these findings establish that, for humans, language and visual concepts are deeply entangled yet the underlying concept survives a change of language. Our work asks the machine analogue of this question, namely whether a VLM's internal visual-textual associations are likewise preserved when the language of the input changes.
 
\paragraph{In-context binding in language and vision-language models.}
Associating an attribute with the correct entity rather than a competing one is a prerequisite for in-context reasoning. \citet{feng2024how} formalised this as the \emph{Binding ID} mechanism, showing through causal interventions that language models tag co-referring entity and attribute tokens with a shared latent identifier occupying a low-rank subspace of the residual stream. Subsequent work has refined and extended this picture, localising an ordering component that determines binding behaviour \citep{dai2024representational}, decomposing retrieval into distinct mechanisms the model mixes as contexts grow complex \citep{feng2025mixing}, recovering the underlying circuitry through learned component masks \citep{davies2023discovering, prakash2024finetuning}, and relating it to the broader linearity of relation decoding and attribute representation in transformers \citep{hernandez2024linearity, heinzerling2024monotonic}. \citet{darshanavlm} extended the Binding ID analysis from text to vision, demonstrating on a synthetic Shapes task that VLMs assign a common identifier to an object's image tokens and its textual mentions. We adopt their task and representative open-weight model as our starting point, and use the same interchange-intervention methodology \citep{meng2022locating, vig2020causal} to locate where binding is computed. Where all of this work establishes binding within a single (English) language, we vary the context and query language independently to test whether the binding is language-invariant. Part of the fragility we uncover originates before the decoder, in tokenizer unfairness toward non-Latin and morphologically rich scripts \citep{petrov2023token_unfairness, rust-etal-2021-good}, and part inside it, which our layer-wise interventions separate.

\section{Linguistic Choices in Detail}
\label{app:lang_details}

\subsection{Cross-family: Maximal Diversity}

Our four cross-family languages were chosen to differ on every major typological axis.
We characterize diversity using the World Atlas of Language Structures (WALS; \citealt{haspelmath2009typological}), a database of structural properties of languages compiled from descriptive grammars, covering features such as canonical word order, morphological type, writing system, and segmentation conventions.
\cref{fig:typology} shows how each cross-family language is classified along four WALS dimensions.
Cells are colored by whether the value is shared with the majority of the set (blue) or divergent (orange); no two languages share the same profile across all four axes.

\definecolor{LightBlue}{HTML}{E8F0FE}
\definecolor{LightYellow}{HTML}{FFF3E0}
\begin{figure}[h]
    \centering
    \includegraphics[width=1\linewidth]{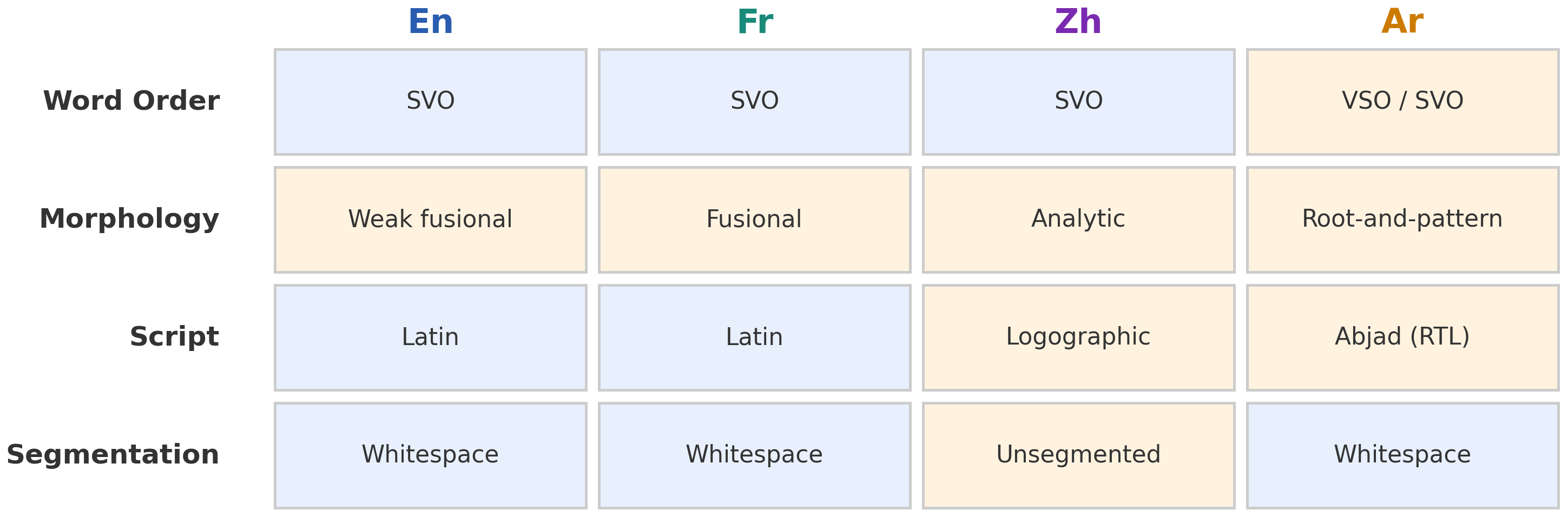}
    \caption{\textbf{Typological profile of the four cross-family languages along WALS dimensions.} Each cell shows the categorical classification; traits are either \colorbox{LightBlue}{shared} with the majority or \colorbox{LightYellow}{divergent}. No two languages match on all four axes.}
    \label{fig:typology}
\end{figure}

\EN{} and \FR{} share SVO (subject-verb-object) order, Latin script, and whitespace segmentation, but differ in morphological type (English has largely lost its inflectional system; French retains richer agreement and gender).
\ZH{} shares SVO order with \EN{}/\FR{} but diverges on every other axis: it is analytic (no inflection), logographic (each character maps to a morpheme), and unsegmented (no whitespace between words).
\AR{} diverges most broadly: it permits VSO order, uses root-and-pattern morphology where three-consonant roots are modified by vocalic templates, writes in a right-to-left abjad, yet segments with whitespace like \EN{}/\FR{}.
These four languages belong to four separate families: Germanic and Romance (both Indo-European), Sinitic (Sino-Tibetan), and Semitic (Afro-Asiatic).
Typological distance vectors from \texttt{lang2vec} \citep{littell2017uriel} confirm near-maximal dispersion in the joint syntactic--phonological--genetic feature space.
A speaker of one cross-family language cannot partially comprehend another without formal training \citep{ringbom2009cross}, unlike within-family pairs where shared vocabulary and structure enable partial comprehension \citep{gooskens2018mutual}.

\subsection{Within-family: Controlled Similarity Gradient}

To disentangle genealogical proximity from script difference, we add two clusters where all languages share the Latin script but vary in closeness to their anchor language.

\paragraph{Germanic (\EN{}, \NL{}, \DE{}).}
All three descend from Proto-Germanic origins\citep{ringe2017protogermanic}.
\NL{} and \EN{} share the closer Ingvaeonic (North Sea Germanic) subgrouping, while \DE{} underwent the High German consonant shift that systematically altered its stop consonants (e.g., English \textit{water} vs.\ German \textit{Wasser}).
\DE{} retains a four-case system, three grammatical genders, and verb-final order in subordinate clauses --- all absent in modern \EN{}.
\NL{} occupies an intermediate position: it preserves two grammatical genders but has largely shed case marking and shares SVO order with \EN{}.
Lexical similarity reflects this ordering: \EN{}--\NL{} ${\sim}$75\%, \NL{}--\DE{} ${\sim}$75\%, \EN{}--\DE{} ${\sim}$60\% \citep{Campbell_Grondona_2008}.
\citet{gooskens2018mutual} confirm that \NL{} speakers can partially comprehend written \EN{} without instruction, while \EN{}--\DE{} intelligibility is measurably lower.

\paragraph{Romance (\FR{}, \IT{}, \ES{}).}
All three descend from Vulgar Latin \citep{posner1996romance}.
\IT{} is generally considered the most conservative major Romance language, retaining the greatest lexical and morphological continuity with the common ancestor.
\FR{}--\IT{} lexical similarity (${\sim}$89\%) is the highest pair in our study, while \FR{}--\ES{} (${\sim}$75\%) is lower due to divergences such as the Latin \textit{f-}~$\to$~\ES{} \textit{h-} shift and richer verb inflection in \ES{} \citep{Campbell_Grondona_2008}.
Untrained readers can often extract the gist of a text in a related Romance language from shared Latinate vocabulary alone \citep{gooskens2018mutual}.

\subsection{Lexical Similarity Structure}

\begin{figure}[h]
    \centering
    \includegraphics[width=0.88\linewidth]{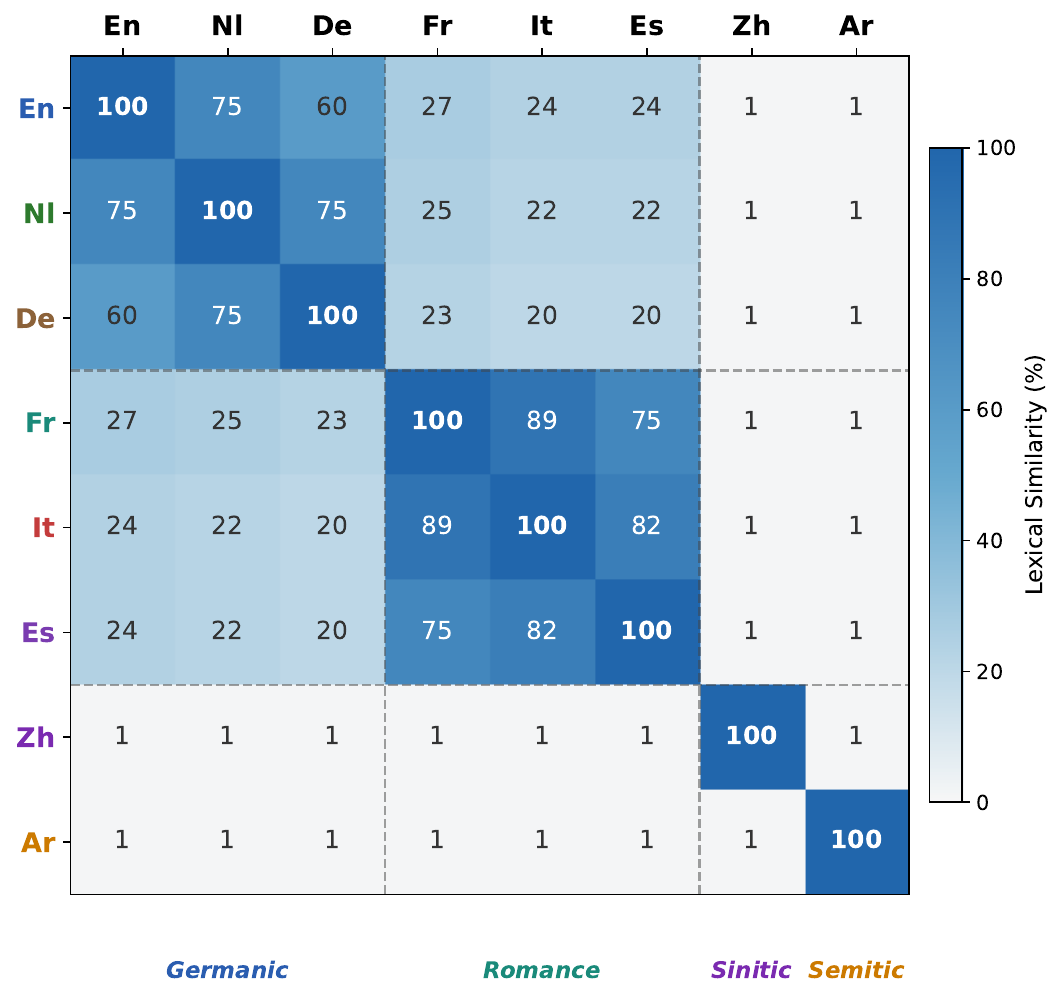}
    \caption{\textbf{Lexical similarity (\%) across all eight languages in M$^{2}$BIND}. Dashed lines separate language families.}
    \label{fig:lexsim}
\end{figure}

\cref{fig:lexsim} shows the full lexical similarity matrix across all eight M$^{2}$BIND languages.
The block-diagonal structure is immediately apparent: Germanic languages form one high-similarity cluster (60--75\%), Romance languages form another (75--89\%), and both \ZH{} and \AR{} show near-zero overlap (${\sim}$1\%) with every other language.
The off-diagonal blocks between Germanic and Romance show modest similarities (20--27\%), reflecting their shared but distant Indo-European ancestry.
While \EN{} and \FR{} have low lexical similarity (27\%), extensive Norman--French borrowing into English gives them a shared Latin-script vocabulary that exceeds what the raw percentage suggests, which may explain the relatively high FM (5.20) for this pair despite the cross-family classification.

\subsection{Connection to Multilingual Representations}

Genealogically, six of our eight languages belong to Indo-European: \EN{}, \NL{}, and \DE{} descend from Proto-Germanic \citep{ringe2017protogermanic}, while \FR{}, \IT{}, and \ES{} descend from Vulgar Latin \citep{posner1996romance}. \ZH{} belongs to Sino-Tibetan and \AR{} to Afro-Asiatic, placing them in entirely separate families with no shared ancestry.
Work on multilingual transformers has shown that models trained on multilingual data develop internal representations where typologically related languages cluster together \citep{pires-etal-2019-multilingual}.
\citet{conneau-etal-2020-unsupervised} demonstrated that cross-lingual transfer improves with shared vocabulary and model scale, while \citet{lauscher-etal-2020-zero} showed that transfer quality degrades with increasing typological distance.
These findings predict exactly the gradient visible in \cref{fig:lexsim}: if the VLM's embedding space reflects linguistic distance, within-family pairs should share sufficiently aligned representations to support binding transfer at the same decoder depth, while cross-family pairs should require additional late-layer reconciliation.
Our ICE analysis (Fig.~2 in the main paper) confirms this at the mechanistic level.

\section{Additional results on Qwen2.5}
\label{app:qwen}

Our main experiments use LLaVA-1.5-OV-7B, whose decoder is itself a Qwen2 model. To test whether the binding behaviour we report depends on that particular stack, we repeat the cross-family protocol on Qwen2.5-VL-7B \citep{qwen25vl}, a separately trained open-weight VLM with its own vision encoder and tokenizer. The images, scenes, gold items $\iota_\star$, and the 16 context--query conditions are identical to those in \cref{sec:results}, and we reuse the same sequence score and Factorization Margin (FM).
\paragraph{Cross-family binding.}
\cref{tab:qwen-pairwise} reports accuracy and FM for Qwen2.5-VL-7B. Two differences from \cref{tab:pairwise} stand out. First, Qwen2.5-VL-7B is a stronger associator in absolute terms. Accuracy is at ceiling (1.000) across the \EN{} and \ZH{} context rows and almost everywhere else, and FM exceeds the corresponding LLaVA cell in all but the two non-Latin monolingual conditions, reaching 6.67 for \EN{}$\to$\AR{}. Binding therefore does not collapse on this model in the way it does on LLaVA, and the headline accuracy gives little indication of any residual fragility. Second, the fragility that remains is visible only through the finer-grained signals. Accuracy dips below ceiling only in the \AR{}-context row (0.950 in its three cross-lingual cells) and in \FR{}$\to$\ZH{} (0.975), which points to a non-Latin language, especially on the context side, as the setting that disturbs top-1 selection. In FM, the weakest binding is the monolingual \ZH{} diagonal (3.71) together with the \AR{}$\to$\ZH{} cross-script cell (3.76), so the two non-Latin scripts again mark the lower end of the binding-strength range even though their accuracy stays high.

\paragraph{Comparison with the main model.}
The agreement between the two models is qualitative rather than numerical. Both place the floor of binding strength on the non-Latin scripts, \ZH{} and \AR{}, and both leave the \EN{}--\FR{} region untouched. The contrast is that LLaVA expresses this fragility as a large FM collapse and an accuracy drop toward 0.90, whereas Qwen2.5-VL-7B absorbs most of it into FM while holding accuracy near ceiling, with the largest accuracy cost appearing under \AR{} context. This supports our central claim in two ways. The non-Latin scripts are the consistent locus of difficulty across independently trained VLMs, and accuracy alone is an unreliable indicator of binding fidelity, since on Qwen a reader of the accuracy column would conclude that multilingual binding is solved while FM shows the \ZH{} and cross-script settings remain measurably weaker.

% ---------------------------------------------------------------
% Qwen2.5-VL-7B cross-family results (Acc. / FM), measured values.
% ---------------------------------------------------------------
\begin{table}[t]
\centering
\footnotesize
\setlength{\tabcolsep}{2.0pt}
\renewcommand{\arraystretch}{1.1}
\begin{tabular}{@{}l*{4}{>{\centering\arraybackslash}p{0.15\columnwidth}}@{}}
\toprule
\textbf{L$_{\mathrm{ctx}}$ $\downarrow$ / L$_{\mathrm{q}}$ $\rightarrow$} &
\textcolor{blue!70!black}{\textbf{En}} &
\textcolor{teal!70!black}{\textbf{Fr}} &
\textcolor{purple!70!black}{\textbf{Zh}} &
\textcolor{orange!80!black}{\textbf{Ar}} \\
\midrule
\textcolor{blue!70!black}{\textbf{En}}   & 1.000 / 5.85 & 1.000 / 6.15 & 1.000 / 4.98 & 1.000 / 6.67 \\
\textcolor{teal!70!black}{\textbf{Fr}}   & 1.000 / 5.88 & 1.000 / 6.00 & 0.975 / 4.60 & 1.000 / 6.15 \\
\textcolor{purple!70!black}{\textbf{Zh}} & 1.000 / 5.31 & 1.000 / 5.80 & \textcolor{red}{1.000 / 3.71} & 1.000 / 4.85 \\
\textcolor{orange!80!black}{\textbf{Ar}} & 0.950 / 4.79 & 0.950 / 4.52 & \textcolor{red}{0.950 / 3.76} & \textcolor{red}{0.975 / 4.43} \\
\bottomrule
\end{tabular}
\vspace{2pt}
\caption{\textbf{Cross-family binding for Qwen2.5-VL-7B.} Rows are context (ctx), columns are query (q). Each cell reports \emph{Acc.\ / FM}. For both, higher ($\uparrow$) is better. Worst cases are \textcolor{red}{highlighted}.}
\label{tab:qwen-pairwise}
\end{table}

\section{Template choices in prompting}
\label{app:prompt}

\begin{figure}[t]
    \centering
    \includegraphics[width=1\linewidth]{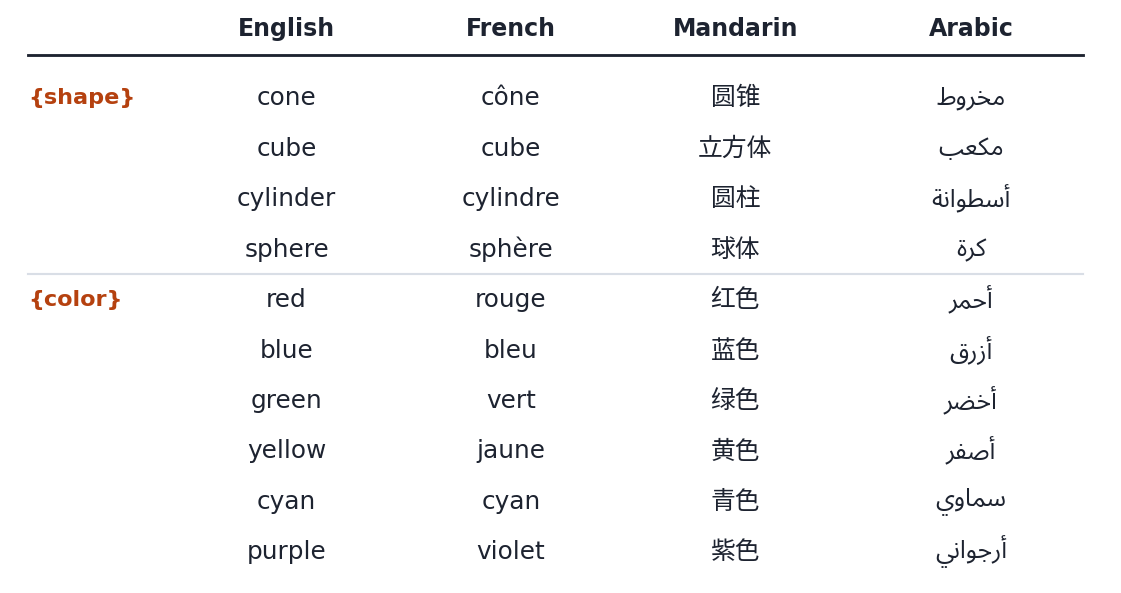}
    \caption{\textbf{Choices in template.} Colour and shape terms that fill the \texttt{\{color\}} and \texttt{\{shape\}} slots of the context and query, shown for the four cross-family languages (\EN{}, \FR{}, \ZH{}, \AR{}). The image and the gold item are held fixed across languages; only these surface forms change.}
    \label{fig:prompt}
\end{figure}

Each instance is a single prompt built from three parts, all derived from the same scene and illustrated earlier in \cref{fig:teaser}. The \textbf{context} introduces every object by its colour and assigns it an item symbol, using one sentence per object. The \textbf{query} names a target object by its shape and asks which item it contains. The \textbf{answer prefix} restates the queried shape and ends just before the item, so that the model's next-token distribution over the candidate symbols is exactly the binding we score. The colour and shape words that fill these slots are listed for every cross-family language in \cref{fig:prompt}.

For example, the scene in \cref{fig:teaser} pairs a cyan cube with item P and a yellow cone with item I, and queries the cone. In \EN{} this yields the context ``The cyan object contains item P. The yellow object contains item I.'', the query ``Which item does the cone contain?'', and the answer prefix ``Answer: The cone contains item \rule{0.8em}{0.4pt}'', whose correct completion is I. The same scene rendered with $\ell_{\mathrm{ctx}}{=}\FR{}$ and $\ell_{\mathrm{q}}{=}\EN{}$ keeps the image and the gold answer fixed while swapping only the surface forms, for example the colour \emph{cyan} becomes \emph{cyan} and \emph{yellow} becomes \emph{jaune} in the context, isolating the effect of language on binding.

\bibliography{arxiv}

\end{document}